\pdfoutput=1

\documentclass[11pt]{article}

\usepackage{acl}

\usepackage{url}
\usepackage[T1]{fontenc}
\usepackage{graphicx}
\usepackage{times}
\usepackage{latexsym}
\usepackage{amsmath}
\usepackage{longtable}
\usepackage{booktabs}
\usepackage[utf8]{inputenc}
\usepackage[T1]{fontenc}
\usepackage[scaled=0.8]{beramono}
\usepackage{microtype}
\usepackage[normalem]{ulem}
\usepackage{enumitem}
\usepackage{inconsolata}
\usepackage{xcolor}
\usepackage{xspace}
\usepackage{CJKutf8} 
\usepackage{multirow}

\newcommand{\sref}[1]{\S\ref{#1}}

\usepackage{tikz}
\usetikzlibrary{calc}
\usetikzlibrary{decorations.pathmorphing}
\usetikzlibrary{shapes,arrows,chains}
\usetikzlibrary{shapes.geometric}
\usetikzlibrary{patterns}
\usetikzlibrary{positioning}
\tikzset{>=latex}
\usepackage{pgfplots}
\usepackage{relsize}
\usepackage{comment}
\usepackage{amssymb}
\usepackage{float}

\DeclareSymbolFont{extraup}{U}{zavm}{m}{n}
\DeclareMathSymbol{\varheart}{\mathalpha}{extraup}{86}
\DeclareMathSymbol{\vardiamond}{\mathalpha}{extraup}{87}

\title{Extracting Lexical Features from Dialects via \\Interpretable Dialect Classifiers}

\author{ Roy Xie$^{\clubsuit}$$^\vardiamond$\quad 
         Orevaoghene Ahia$^{\spadesuit}$ \quad
         Yulia Tsvetkov$^{\spadesuit}$ \quad 
         Antonios Anastasopoulos$^{\vardiamond}$ \\
         $^\clubsuit$ Duke University \\
         $^\spadesuit$ Paul G. Allen School of Computer Science \& Engineering, University of Washington \\
         $^\vardiamond$ Department of Computer Science, George Mason University \quad \\
         \texttt{ruoyu.xie@duke.edu\quad \{oahia, yuliats\}@cs.washington.edu\quad antonis@gmu.edu}\\
    }

\begin{document}
\maketitle

\begin{abstract}
Identifying linguistic differences between dialects of a language often requires expert knowledge and meticulous human analysis. This is largely due to the complexity and nuance involved in studying various dialects. 
We present a novel approach to extract distinguishing lexical features of dialects by utilizing interpretable dialect classifiers, even in the absence of human experts. We explore both post-hoc and intrinsic approaches to interpretability, conduct experiments on Mandarin, Italian, and Low Saxon, and experimentally demonstrate that our method successfully identifies key language-specific lexical features that contribute to dialectal variations.%
\footnote{Data and code are available: \url{https://github.com/ruoyuxie/interpretable_dialect_classifier}}
\end{abstract}

\section{Introduction}
Dialects and closely related languages exhibit subtle but significant variations, reflecting regional, social, and cultural differences \cite{chambers1998dialectology}. Identifying and distinguishing differences between these dialects is of great importance in linguistics, language preservation, and natural language processing (NLP) research \citep{salameh-etal-2018-fine,goswami-etal-2020-unsupervised}. Traditionally, identifying the specific linguistic features that distinguish dialects of a language has relied on manual analysis and expert knowledge \citep{cotterell-callison-burch-2014-multi}, as the differences between these dialects could be subtle and hard to detect without linguistic expertise \citep{zaidan-callison-burch-2011-arabic}. This process is also time-consuming and is usually language-specific due to the peculiarities that different languages exhibit. 

Extracting distinctive words of particular dialects is essential for studying dialectal variation, especially in dialectology \citep{chambers1980dialectology}. In this work, we focus on extracting word-level distinguishing and salient features in dialects, also called `shibboleths' \cite{prokic2012detecting}. We propose an automated workflow that could potentially assist researchers such as dialectologists and corpus linguists in their analysis of dialectal variations. To achieve this, we leverage strong neural classifiers of dialects \citep{aepli-etal-2023-findings,scherrer2023tenth,scherrer2022proceedings} and pair them with model interpretability techniques to extract these features.

We hypothesize that there are certain distinguishing features in dialects that the models learn during training, which enables them to make accurate predictions at test time. We utilize post-hoc \citep{Simonyan2014DeepIC,DBLP:journals/corr/RibeiroSG16} and intrinsic \citep{10.5555/3327757.3327875,DBLP:journals/jmlr/ArikP20} feature attribution explanation methods to extract these features from model interpretations, in the form of local explanations from dialect classifiers. 

\begin{figure}[t]
\small

    \centering  
    \vspace{-1em}
    \includegraphics[width=7.5cm]{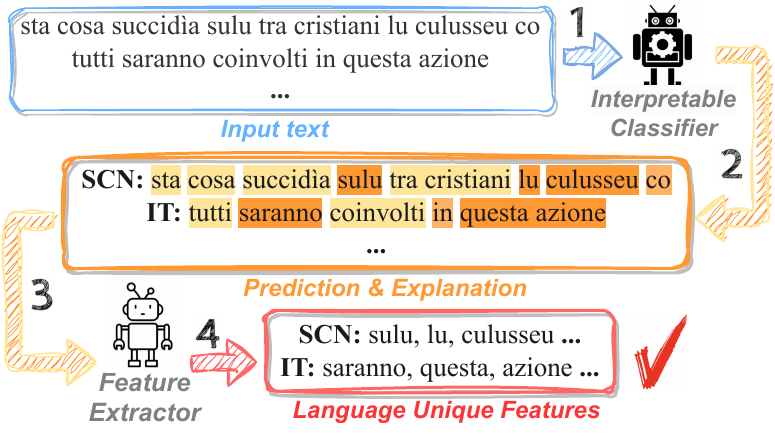}
    \caption{\textit{(1)} given an input text; \textit{(2)} the interpretable dialect classifier return labels (SCN and IT) and explanations; \textit{(3)} the extractor takes the explanations and \textit{(4)} outputs meaningful features to the languages.}
    \label{fig:workflow}
    \vspace{-1.5em}
\end{figure}

Our experiments focus on three language groups: Mandarin, Italian, Low Saxon, and their respective dialects. We demonstrate the effectiveness of our approach through automatic evaluation, human evaluation, and extensive analysis. We use known lexical features of some of the dialects we consider and show the viability of using explanation methods to extract unique dialectal features.

\section{Background and Related Work}
The importance of interpreting model decisions has increasingly been recognized in recently \citep{belinkov-etal-2020-interpretability,DBLP:conf/icml/KohL17,mosca-etal-2022-shap,rajagopal-etal-2021-selfexplain}, primarily due to its role in deciphering the inner workings of black-box models. Two main approaches include: \textit{(i)} post-hoc methods \cite{DBLP:conf/nips/LundbergL17, DBLP:journals/corr/RibeiroSG16, DBLP:journals/corr/abs-1911-06194} which provide insights into the predictions of pre-existing models based on model internals; \textit{(ii)} intrinsic methods \cite{rajagopal-etal-2021-selfexplain,10.5555/3327757.3327875, rigotti2021attention} where interpretability is an integral feature that is optimized concurrently with the model's main primary task during training.  

This paper also closely relates to dialect classification \citep{aepli-etal-2023-findings,scherrer2023tenth,scherrer2022proceedings,jauhiainen2019automatic}. Some high-performing neural dialect classifiers are able to achieve $>$ 90\% accuracy even when there are multiple categories and levels of noise present or there are just very subtle differences in general \citep{srivastava-chiang-2023-fine}. Most languages worldwide are under-resourced \citep{joshi-etal-2020-state}, and basic tasks like identifying verbs and nouns within a dialect can be challenging, given that syntactic parsing is primarily efficient for well-resourced languages \cite{hellwig2023data,hou2022syntax}. In this work, we focus on extracting lexical differences between dialects, as they show potential in distinguishing dialectal variations. The idea of automatically extracting linguistic features is not new \citep{brill1991discovering,demszky-etal-2021-learning}. However, we identify these features \textit{through the lens of model explanations} by using feature-attribution methods. To the best of our knowledge, this is the first study to undertake such an endeavor.

\section{Method}
We present a simple yet effective method to extract lexical differences that distinguish dialects using explanations obtained from interpretable dialect classifiers as shown in Figure \ref{fig:workflow}.

\subsection{Interpretable Dialect Classifier} 
Our interpretable dialect classifier, built on top of transformer-based models, is designed to work with both post-hoc and intrinsic interpretable methods. For the scope of this work we focus on Leave-One-Out (LOO), a popular model-agnostic feature-attribution method, and SelfExplain \citep{rajagopal-etal-2021-selfexplain}, an intrinsic interpretable method that learns to attribute text classification decisions to relevant parts in the input text. For the intrinsic approach, we incorporate the model encoder into the SelfExplain framework, exclusively extracting only local explanations. For the  post-hoc approach, we train a separate classifier to calculate the LOO interpretations. 

\subsection{Explanation Methods}\label{appendix:explanation_methods}

We start with a dialect classifier trained to take an input sentence $X$ and predict its corresponding label $\mathbf{y}$. Let $\mathbf{u}_s$ be the final layer representation of the ``[CLS]'' token for $\mathbf{X}$, which is the sentence representation typically used to make a prediction.

\paragraph{Post-hoc Approach}
During inference, LOO estimates the attribution score of each token ${x_i}$ in input $\mathbf{X}$ in relation to model's prediction $\mathbf{\hat{y}}$. To do so, $\mathbf{u}_s$  is passed through ReLU, affine, and softmax layers to yield a probability distribution over outputs. For each feature ${x_i}$, LOO calculates the change in probability when $\{x_i\}$ is removed from $\mathbf{X}$. Let $X \textbackslash \{x_i\}$ denote input $\mathbf{X}$ without feature ${x_i}$ and $u_i$ the final layer representation of the ``[CLS]'' token for $X \textbackslash \{x_i\}$ . We term this the relevance score and expect that influential features/explanations in the input $\mathbf{X}$ will have higher scores. 
\begin{align}
\boldsymbol{\ell} &= \operatorname{softmax}(\mathrm{affine}(\mathrm{ReLU}(\mathbf{u}_s))) \label{eq:bh}
\end{align}
\vspace{-2.5em}
\begin{align*}
\boldsymbol{\ell_i} &= \operatorname{softmax}(\mathrm{affine}(\mathrm{ReLU}(\mathbf{u_i})))\\ 
\boldsymbol{\mathcal{r}_i} &= \boldsymbol{\ell} - \boldsymbol{\ell_i}
\end{align*}

\paragraph{Intrinsic Approach}
For our intrinsic approach using SelfExplain \citep{rajagopal-etal-2021-selfexplain}, we augment the dialect classifier with a Local Interpretability Layer (LIL) during training. This layer quantifies the relevance of each feature ${x_i}$ in input $\mathbf{X}$ to the final label distribution $\boldsymbol{\ell}$ via activation difference \citep{10.5555/3305890.3306006}, and is trained jointly with the final classifier layer. Taking $\boldsymbol{\ell}$ in \autoref{eq:bh}, the loss is the negative log probability, summed over all training instances:
\begin{align*}
L_{\mathit{dialect-classifier}} &= -\textstyle\sum_i \log \boldsymbol{\ell}[y_i^\ast]
\end{align*}

where $y^*_i$ is the correct label for instance $i$. To obtain the attribution score of each feature ${x_i}$ in input $\mathbf{X}$, we
first estimate the output label distribution without ${x_i}$ by transforming the difference between $\mathbf{u}_s$ and $\mathbf{u}_j$, where $\mathbf{u}_j$ is the MLM representation of feature ${x_i}$:
\begin{align*}
 \mathbf{s}_j &= \mathrm{softmax}(\mathrm{affine}(\mathrm{ReLU}(\mathbf{u}_s) - \mathrm{ReLU}(\mathbf{u}_j)))
 \end{align*}\begin{align*} 
\mathrm{loss} &= L_{\mathit{dialect-classifier}} + \alpha_1 L_{\mathit{LIL}}  \label{eq:combined_loss}
\end{align*}

 The relevance of each feature ${x_i}$ can be defined as the change in probability of the correct label when ${x_i}$  is included vs.~excluded: 
\begin{align*} 
r_j & = [\boldsymbol{\ell} ]_{y^*_i}- [\mathbf{s}_j ]_{y^*_i} 
\end{align*}

where higher $r_j$ signifies more relevant features to the prediction, serving as better explanations.

\paragraph{Mapping Explanations to Lexical Features } \label{extration_method}
We extract explanations from the classifiers outlined above. Note, however, that these explanations are at the \textit{sentence} level, but one ideally would need features that in general identify/describe one dialect in contrast to another at the \textit{language} level, i.e., at the \textit{corpus} level. To achieve this, we devise a corpus-level feature extraction method that takes sentence-level explanations as input and produces ``global'' features.\footnote{In this study, we extract distinguishing lexemes (words) through unigrams. Nonetheless, our approach can be readily adapted to phrase-level analysis using ngrams, though such an extension falls beyond the purview of this work.}

Given a set of sentence-level explanations $E = \{e_1, e_2, \ldots, e_n\}$ from a classifier, we first filter out explanations from incorrect predictions or those that are not unique to a specific language variety. Let $E'$ represent the filtered set of explanations: 
\begin{equation*}
  E' = \{ e \in E \mid \text{isCorrect}(e) \land \text{isUnique}(e) \}
\end{equation*}

Next, we apply Term Frequency-Inverse Document Frequency (TF-IDF) to $E'$ to extract the most salient global features. Let $F$ be the final set of extracted features. The TF-IDF score for a term $t$ in a document $d$ in a corpus $D$ is given by:
\begin{equation*}
  \text{TF-IDF}(t, d, D) = \text{TF}(t, d) \times \text{IDF}(t, D)
\end{equation*}
We can then define the feature extraction as:
\begin{equation*}
  F = \{\text{TF-IDF}(t, d, E') \mid t \in d, d \in E'\}
\end{equation*}

\section{Languages and Datasets} 
\label{appendix:languages_and_datasets}
\subsection{Languages}
We discuss below each of the language continua, which have multiple dialects that vary distinctly across different regions. We study three distinct language continua and their respective dialects: Mandarin, Italian, and Low Saxon (Dutch and German). Our selection is largely influenced by cultural and typological diversity concerns, but also by the dearth of dialectal data for other languages. More information about the dialects can be found in Appendix~\ref{appendix:dialects_info}.

\subsection{Datasets}
We use FRMT dataset~\cite{DBLP:journals/corr/abs-2210-00193} for Mandarin dialects. FRMT provides sentence- and word-level translations of English Wikipedia text into Mainland Mandarin (CN) and Taiwan Mandarin (TW). 
For Italian dialects, we combine data from the Identification of Languages and Dialects of Italy (ITDI) shared task \cite{aepli-etal-2022-findings} and Europarl v8 corpus \cite{koehn-2005-europarl}. ITDI provides 11 Italian dialects obtained from crawling Wikipedia, and we use them to mix with the same amount of data from standard Italian in Europarl. 
For Low Saxon dialects, we use LSDC dataset \cite{siewert-etal-2020-lsdc}, which consists of 16 West-Germanic Low Saxon dialects from Germany and the Netherlands. 
Appendix \ref{appendix:all_dataset} shows the statistics for all datasets. We remove all punctuations and lowercase all Latin characters to focus on extracting lexical features.

\section{Experiments}
In this section, we present multiple experiments
and demonstrate that our method results in distinguishing and meaningful feature extraction. For the main results, we focus on Mandarin (CN-TW) and Sicilian-Italian (SCN-IT). A total of four annotators are involved in the evaluation process.\footnote{All annotators are proficient in the annotated language pairs. Three annotators work on CN-TW, and one for SCN-IT, as it is difficult to find annotators who are proficient in multiple dialects.} 

\subsection{Models}
Our dialect classifiers are built on XLM-RoBERTa base \cite{conneau-etal-2020-unsupervised}. We maintain the hyperparameters and weights from the pre-training of the encoders and train the models for 5 epochs with a batch size 16 for each language pair. All experiments are done on an NVIDIA A100 GPU. 
 
\subsection{TF-IDF Baseline}\label{baseline}
As a baseline, we use TF-IDF to evaluate how effectively it extracts meaningful lexical features from CN-TW language pair (Table \ref{table:baseline_result}). We present annotators with the extracted lexical features using TF-IDF, asking them to determine whether these features are salient and unique to the language.\footnote{The annotators are given options to select if a feature is likely to belong to one, both, or neither of the dialects.}

\begin{table}[h!]
\centering
\small
\begin{tabular}{@{}ccc@{}}
\toprule
Option (\%) & CN & TW \\
\midrule
CN & \textbf{70.0} & 0.0 \\
TW & 0.0 & \textbf{60.0} \\
Both & 30.0 & 40.0 \\
None & 0.0 & 0.0 \\
\bottomrule
\end{tabular}
\caption{Baseline results for capturing language-unique features for CN-TW.}
\label{table:baseline_result}
\vspace{-1em}
\end{table}

\subsection{Explanation Evaluation} 
We hypothesize that good explanations from highly performing dialect classifiers should be subsets of the input that represent distinctive features of the respective dialect in which the input is written. We first evaluate the general robustness of explanations using \textbf{Sufficiency} \citep{Jacovi2018UnderstandingCN,yu-etal-2019-rethinking} (Do explanations adequately represent the model predictions?)  and \textbf{Plausibility} \cite{ehsan2019automated,hayati-etal-2023-stylex} (Do explanations seem credible and comprehensible to humans?)

\paragraph{Sufficiency}
We train a separate classifier to perform the dialect classification task with only the explanations as input, and 
the predicted labels as target.
Higher accuracy indicates that the explanations are more reflective of the model predictions. We train these models with the top ranking
explanations of each sentence as input, and present the results in Table \ref{table:Sufficiency} for both explanation methods. Both methods achieve over 90\% accuracy when k $\geq$ 3, which sets a reliable baseline for further evaluation. CN-TW and SCN-IT have an average sentence length of 18.8 and 16.2, respectively, which implies that our sufficiency scores are trustworthy, as we obtain them with less than 20-30\% of an average sentence.

\begin{table}[thp!]
  \centering
  \small
  \begin{tabular}{ccccc}
    \toprule
    Methods & Dialects & $k=1$ & $k=3$ & $k=5$ \\
    \midrule
    \multirow{2}{*}{SelfExp} &
    CN-TW      & 76.5 & 96.7 & 97.8 \\
    & SCN-IT     & 87.8 & 95.8 & 97.9 \\
    \midrule

    \multirow{2}{*}{LOO} &
    CN-TW     & 81.3 & 93.2 & 97.2 \\
    & SCN-IT     & 87.4 & 95.4 & 96.6 \\
    \bottomrule
  \end{tabular}
    \caption{More explanations lead to higher sufficiency. Both explanation methods are over 90\% accurate when $k\geq 3$, setting a reliable baseline for future evaluation.}
  \label{table:Sufficiency}
  \vspace{-1em}
\end{table}

\begin{figure}[thp!]
\small

    \centering  
    \includegraphics[width=7.5cm]{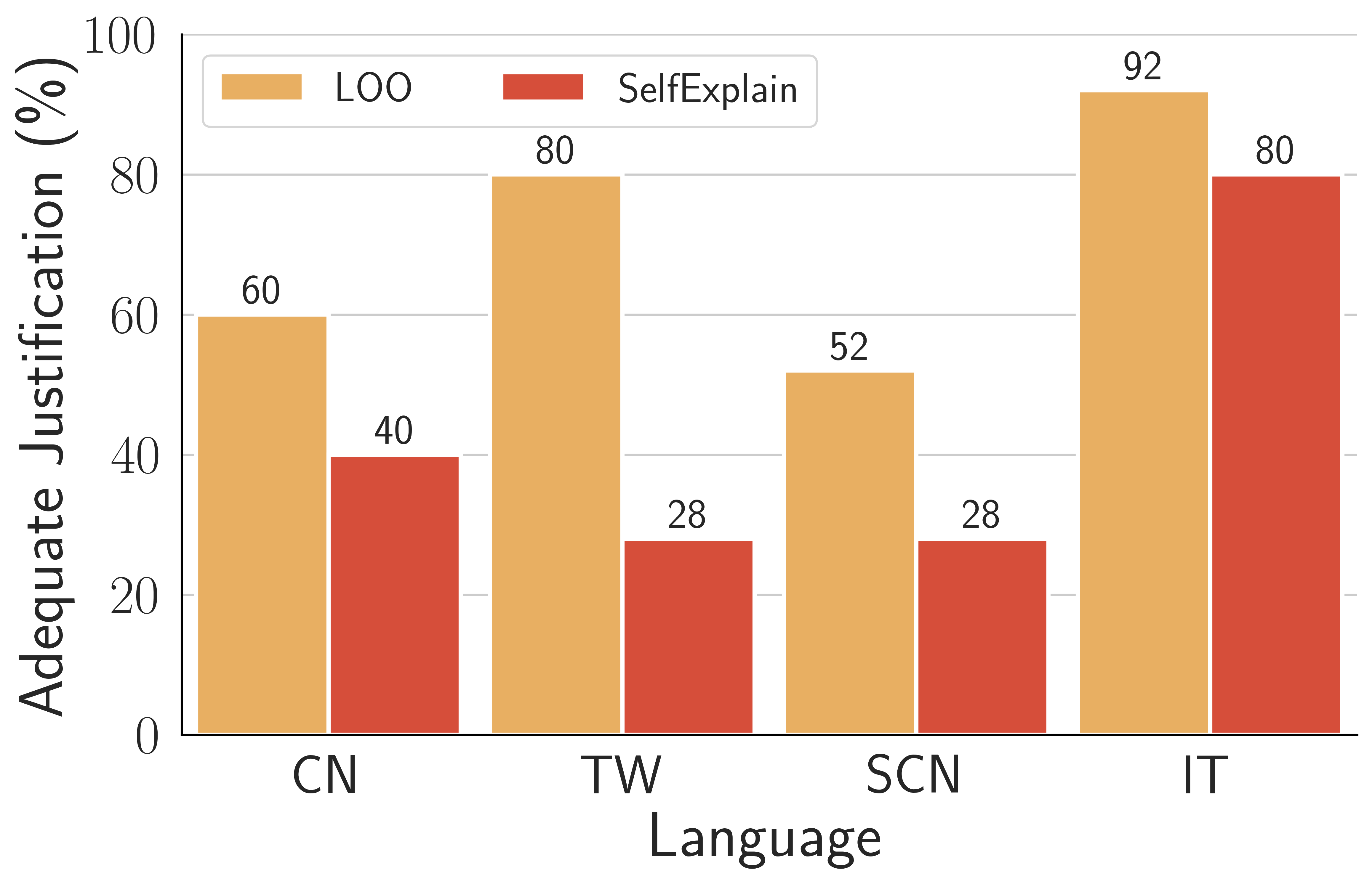}
        \caption{Adequate justification percentage for LOO and SelfExplain. Humans found LOO produces more justifiable explanations across all four dialects.}
    \label{fig:Plausibility}
    \vspace{-1em}
\end{figure}

\paragraph{Plausibility}
We give each annotator 25 sentences with the model's predictions and the top three explanations from LOO and SelfExplain. We randomly shuffle and anonymize the explanation methods and ask annotators \textit{(i)} Should the model classify a given sentence in certain dialects based on the explanations? \textit{(ii)} If the explanations do not adequately justify the model's prediction, what should the model's prediction be based on?
The annotators are given options to select one, both, or neither explanation method. We present the percentage of instances that are adequately justified according to the annotators on Figure~\ref{fig:Plausibility}. Overall, LOO achieves a higher percentage of perceived adequate justification, compared to SelfExplain. Therefore, we will use LOO as the main explanation method for the rest of the studies in this work.

\subsection{Can we extract lexical features by interpreting dialect classifiers?}
\paragraph{Automatic Evaluation} As a sanity check, we propose a simple automatic evaluation metric, \textsc{pick-up rate} (PR), to evaluate the ability of explanation methods on capturing language-unique features. Based on comparative dialectology and other linguistic literature, we identify a small set of distinctive features for the languages we work with.\footnote{After reviewing the literature, we identified features that can be easily measured, which allows us to potentially show that the features that humans described align with the outputs generated by the model.} We define PR as the likelihood of language-unique features being captured in the input sample. The higher the PR is, the better the explanation method aligns with some ground truth features. Given \( e_{g} \) as the number of times that a language-unique feature \( g \) is used as an explanation \( e \) and \( c_{g} \) as the number of times that \( g \) appears in a corpus \( c \), $ PR(g) = \frac{\text{count}(e_{g})}{\text{count}(c_{g})} $. 

FRMT provides a set of Mandarin word translations between CN and TW. We present the PR result for these translated words in Appendix~\ref{appendix:pr_cn-tw}. For both explanation methods, the language-unique translations tend to only appear in their corresponding classes with reasonable pick-up rates. Note that script differences between CN and TW make its fairly easy for the classifiers to correctly classify them. Therefore, we conduct the same experiments on two additional Low Saxon dialect pairs for confirmation (Appendix~\ref{appendix:low_saxon_pr}). All experiments showed similar patterns, indicating that language-unique lexical features can indeed be retrieved from the explanations. Further discussion about PR can be found in Appendix \ref{appendix:pr_discussion} and \ref{appendix:exp_from_train}.
\label{sec:pr_result}

\paragraph{Human Evaluation}
\label{subsec:without}
Similar to the baseline in \sref{baseline}, we select the top 20 extracted features using our extraction method as described in \sref{extration_method}. The inter-annotator agreement statistics can be found on Appendix \ref{appendix:agreement_statistics}. We calculate the percentage of choices made in the each option relative to the total choices the and present the CN-TW and SCN-IT results in Table~\ref{table:main_result} and compare CN-TW results with baseline in Table~\ref{table:combined_result}.

\begin{table}[h!]
  \centering
 \small
  \begin{tabular}{@{}c|cc|c|cc@{}}
    \toprule
    Option (\%) & CN  & TW & Option (\%) & SCN & IT \\
    \midrule
    CN & \textbf{88.9} & 10.5 & SCN & \textbf{45.0} & 0.0 \\
    TW & 0.0 & \textbf{89.5} & IT & 15.0 & \textbf{63.2} \\
    Both & 11.1 & 0.0 & Both & 40.0 & 36.8 \\
    None & 0.0 & 0.0 & None & 0.0 & 0.0  \\
    \bottomrule
  \end{tabular}
  \caption{Our extraction method effectively extracts language-unique features in two language pairs, CN-TW and SCN-TW.}
  \label{table:main_result}
\vspace{-1em}

\end{table}

We observe that for CN-TW, which is easily distinguishable, our method captures 88.9\% and 89.5\% of CN and TW features, respectively. Compare to the baseline in \sref{baseline} (70\% and 60\%), our method is \textbf{27\%} and \textbf{49\%} higher. The numbers for SCN and IT are slightly lower (45\% for SCN and 63.2\% for IT) but it is important to note that the two languages do share a large percentage of their vocabulary, so we believe that these scores are indeed encouraging. The reason is that the method is confirmed to be rather precise: None of the suggested features for Italian would be appropriate for Sicilian, and only 15\% of the suggested SCN features would not be appropriate for it.

\begin{table}[h!]
\centering
\small
\begin{tabular}{@{}c|cc|cc@{}}
\toprule
& \multicolumn{2}{c|}{Baseline} & \multicolumn{2}{c}{Ours} \\
Option (\%) & CN & TW & CN & TW  \\
\midrule
CN & 70.0 & 0.0 & \textbf{88.9} & 10.5  \\
TW & 0.0 & 60.0 & 0.0 & \textbf{89.5}  \\
Both & 30.0 & 40.0 & 11.1 & 0.0  \\
None & 0.0 & 0.0 & 0.0 & 0.0  \\
\bottomrule
\end{tabular}
\caption{Comparison of the baseline and our extraction method on capturing language-unique features for CN-TW. Our method significantly outperforms the baseline, capturing nearly \textbf{90\%} of the language-unique features for both languages.}
\label{table:combined_result}
\vspace{-1em}
\end{table}

\subsection{Classification Accuracy} While our primary goal is to extract lexical features, ensuring high classification accuracy is also crucial as incorrect predictions could undermine explanations. We train 21 distinct models for binary classification for all dialect pairs and present their results on Appendix~\ref{appendix:classification_acc_baseline}. We observe our method achieves high accuracy across all language pairs, with an average of 98.7\%. This ensures that the features extracted by the model are supported by reliable predictions, thereby enhancing the value and reliability of the explanations it provides.

\section{Conclusion} 
In this work, we introduce a novel approach for capturing language-unique lexical features from dialects through interpretable dialect classifiers. We utilize both post-hoc and intrinsic explanation methods and experiment on three language groups - Mandarin, Italian, and Low Saxon, and their respective dialects, conducting extensive evaluation and analysis to showcase the effectiveness of our method. In the future, we plan to broaden this approach to address additional linguistic aspect beyond the lexical level (Appendix \ref{appendix:other_features}). 

More broadly, our paper takes a first step to assess how interpretability techniques can be used to unearth lexical and, potentially, other linguistic features. By doing so, we hope to provide a framework that future studies can build upon.

\section{Ethics Statement}
We envision a future where researchers, regardless of their expertise in a particular language or dialect, can leverage our method to gain insights of dialectal variations. While our primary intention is to promote dialectal inclusivity, we realize that, like any tool, it could be misused in ways that might lead to division or stereotyping. As dialects can be associated with specific ethnic groups or nationalities, the technology might be misused for profiling purposes. It might introduce misleading correlations and negatively impact certain groups. 

\section{Limitations}
A limitation to our work is that we are working with binary classification and data that we \textit{a priori} know to belong to a certain language variety, to provide a proof-of-concept. If one was applying this work in the real world, e.g. on data collected from multiple locations within a language's geographical area, we could substitute our classification scheme to now predict the location or class of the data collection. 

While our emphasis is on lexical aspects, it is crucial to acknowledge the broad spectrum of other linguistic elements that contribute to the richness and complexity of dialects, such as syntactic, phonetic, and semantic features. Our method demonstrates excellent performance in lexical feature extraction, it may not yet adeptly identify and analyze these additional facets of linguistic variation. Additionally, finding annotators proficient in multiple dialects is challenging. Therefore, some experiments were only conducted on a subset of languages due to the limited availability of data and annotators. In the future, we plan to extend our method to more languages and integrate modules that will focus specifically on these diverse linguistic features, making it a more comprehensive tool for dialect analysis.

\section{Acknowledgements}
The authors are thankful for the anonymous reviewers for their feedback. This work was generously supported by NSF Award IIS-2125466 as well as through a GMU OSCAR
award for undergraduate research. 
This research is supported in part by the Office of the Director of National Intelligence (ODNI), Intelligence Advanced Research Projects Activity (IARPA), via the HIATUS Program contract \#2022-22072200004. 
This material is also funded by the DARPA Grant under Contract No.~HR001120C0124. 
We also gratefully acknowledge support from NSF CAREER Grant No.~IIS2142739, NSF Grants No.~IIS2125201, IIS2203097, and the Alfred P.~Sloan Foundation Fellowship.
The views and conclusions contained herein are those of the authors and should not be interpreted as necessarily representing the official policies, either expressed or implied, of ODNI, IARPA, or the U.S. Government. The U.S.~Government is authorized to reproduce and distribute reprints for governmental purposes notwithstanding any copyright annotation therein.

\bibliography{anthology,custom}

\clearpage
\newpage
\appendix
\onecolumn

\section{Dialects}\label{appendix:dialects_info}

\begin{figure}[thp!]
    \centering  
    \includegraphics[width=7.5cm]{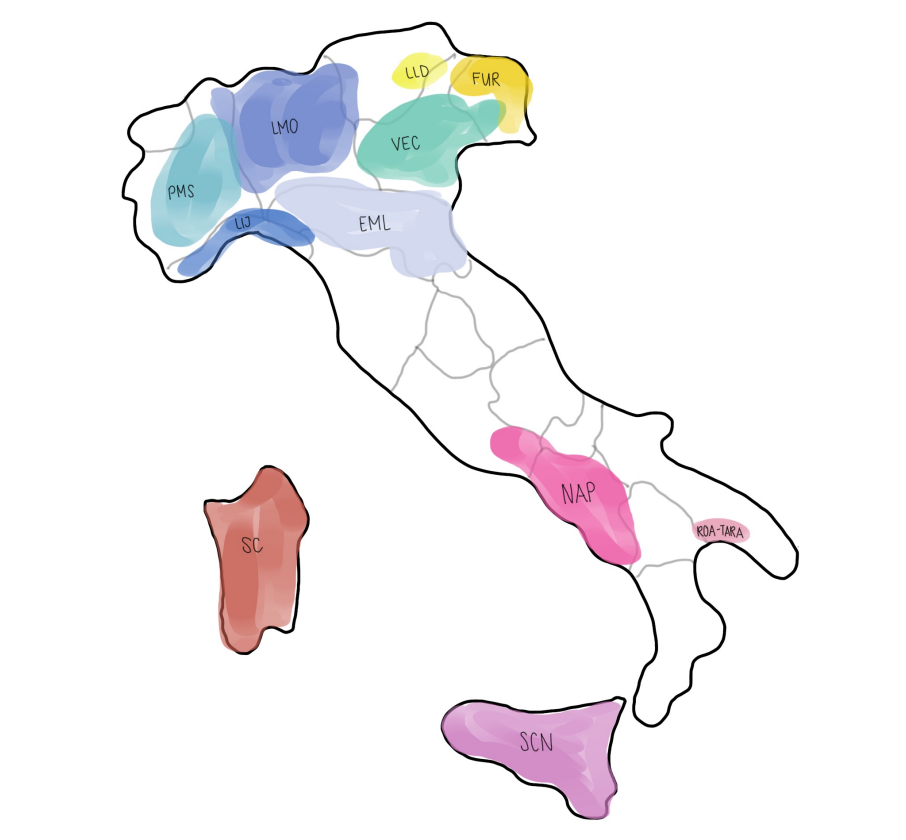}
    \caption{A general overview of the geographical areas in Italy for the 11 languages and dialects. While the map's vague due to the complexity of the situation, it provides a rough idea of where in Italy to locate the varieties. The map is sourced from \citet{aepli-etal-2022-findings}.}

    \vspace{-1em}
    \label{fig:idti_map}
\end{figure}

\begin{figure}[thp!]
    \centering  
    \includegraphics[width=7.5cm]{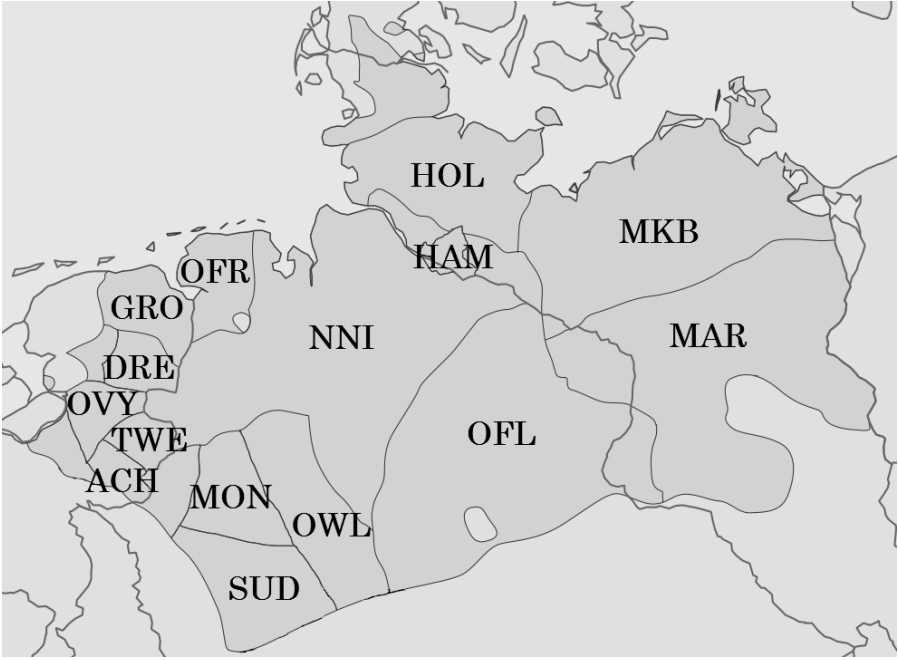}
    \caption{Rough regions where the 16 considered Low Saxon languages and dialects are spoken. This map is taken from \citet{siewert-etal-2020-lsdc}.}

    \label{fig:lsdc_map}
\end{figure}

\paragraph{Mandarin Dialects}
Mandarin, also known as Standard Chinese or Putonghua, is  part of the Sinitic language family and one of the most widely spoken languages worldwide. We focus on two variations of Mandarin: Mainland Mandarin (CN) and Taiwan Mandarin (TW). The two varieties are closely related and they share a core vocabulary, but there are variations in the usage of certain words and phrases. In writing, Mainland Mandarin also has adopted simplified Chinese characters, while Taiwan uses traditional characters. 

\paragraph{Italian Varieties}
Italian, a Romance language, consists of a diverse range of dialects across different regions of Italy. While these dialects have evolved from Latin and share many common words, they exhibit variations in phonetics, vocabulary, and grammar (Figure \ref{fig:idti_map}). For example, Sicilian, spoken in Sicily and the southern regions of Italy, presents distinct phonological features and a rich vocabulary influenced by Arabic and other languages, and as a result it has its own grammar, vocabulary, and pronunciation rules, to the point that it can be difficult for Italian speakers to understand. We note that a lot of the Italian vernaculars are categorized as distinct languages with their own ISO codes (e.g. Venetian, Neapolitan, Sicilian, to name a few). But nevertheless it is undeniable that all of them fall within the same branch of Italic languages and in practice for a diverse language continuum.

\paragraph{Low Saxon Dialects}
Low Saxon is a West Germanic language that encompasses a range of dialects spoken in the northern regions of the Netherlands and Germany (Figure \ref{fig:lsdc_map}). There are multiple dialects that vary distinctly across different regions. For example, the dialects spoken in the Netherlands, such as Gronings and Twents, have distinct vowel sounds and consonant variations compared to the German dialects, such as Plattdeutsch.

\section{Dataset Statistics}\label{appendix:all_dataset}

\vspace{-3mm}
\begin{table*}[h!]
\centering
\begin{tabular}{c|c|c|c}
\toprule
\textbf{Language} & \textbf{Dialect Region} & \textbf{Train} & \textbf{Test} \\
\midrule
\multirow{2}{*}{Mandarin}
& Mainland (CN) & 3802  & 467  \\
& Taiwan (TW) &  3807 & 488  \\
\midrule
\multirow{11}{*}{Italian}
& Piemonte (PMS) & 3305  &  414 \\
& Veneto (VEC) &  11249 &  1447 \\
& Sicilia (SCN) & 3250  &  399 \\
& Campania (NAP) & 2012  & 254  \\
& Emilia Romagna (EML) & 1648  & 222  \\
& Taranto (ROA TARA) & 716  &  90 \\
& Sardegna (SC) & 810  & 99  \\
& Liguria (LIJ) & 4575  &  558 \\
& Friuli (FUR) &  2990 & 368  \\
& Veneto (LID) &  - & -  \\
& Lombardia (LMO) & 5846  & 733  \\
\midrule
\multirow{16}{*}{Low Saxon}
& Achterhoek (ACH) & 791  & 106  \\
& Drenthe (DRE) &  5322 & 634  \\
& Groningen (GRO) & 27  & 2  \\
& Hamburg (HAM) & 5559  &  705 \\
& Holstein (HOL) & 10381  & 1293  \\
& Mark-Brandenburg (MAR) &  177 & 20  \\
& Mecklenburg-Vorpommern (MKB) &  15654 & 1913  \\
& Munsterland (MON) & 589  & 86  \\
& Northern Lower Saxony (NNI) & 649  & 80  \\
& Lower Prussia  (NPR) & 298  &  33 \\
& Eastphalia (OFL) & 7512  &  952 \\
& East Frisia (OFR) & 197  &  25 \\
& Overijssel (OVY) & 1063  & 113  \\
& Eastern Westphalia (OWL) & 11480  & 1396  \\
& Sauerland (SUD) & 13747  & 7425  \\
& Twente (TWE) &  547 &  59 \\
\bottomrule
\end{tabular}
\caption{Dataset Statistics}
\end{table*}

\clearpage
\newpage
\section{PR Results}

\subsection{PR on CN-TW}\label{appendix:pr_cn-tw}
\begin{table}[thp!]
\begin{center}
\begin{CJK*}{UTF8}{} 
\resizebox{0.55\textwidth}{!}{%
\begin{tabular}{cc|cc|cc}  
\toprule
\multicolumn{2}{c}{} & \multicolumn{2}{c}{LOO} & \multicolumn{2}{c}{SelfExp.} \\  
Label & Feature & CN PR & TW PR & CN PR & TW PR\\  
\midrule
\multirow{11}{*}{CN}& \CJKfamily{gbsn}菠萝 & 12.5 & 0 & 12.5 & 0  \\
& \CJKfamily{gbsn}鼠标 & 38.5 & 0 & 15.4 & 0 \\
& \CJKfamily{gbsn}新西兰 & 50 & 0  & 66.7 & 0 \\
& \CJKfamily{gbsn}打印机 & 33.3 & 0 & 33.3 & 0 \\
& \CJKfamily{gbsn}站台 & 0 & 0  & 0 & 0 \\
& \CJKfamily{gbsn}过山车 & 57.1 & 0 &  14.3 & 0 \\
& \CJKfamily{gbsn}三文鱼 & 100 & 0  & 100 & 0 \\
& \CJKfamily{gbsn}洗发水 & 28.6 & 0 &  28.6 & 0 \\
& \CJKfamily{gbsn}软件 & 60 & 0 &  0 & 0 \\
& \CJKfamily{gbsn}悉尼 & 0 & 0 & 50 & 0 \\
& \CJKfamily{gbsn}回形针 & 50 & 0 &  33.3 & 0 \\
\midrule
\multicolumn{2}{c}{\textit{Avg. PR}} &  \textbf{\textit{39.1}} &  \textbf{\textit{0}} &  \textbf{\textit{35}} &  \textbf{\textit{0}}  \\
\midrule
\multirow{11}{*}{TW}& \CJKfamily{bsmi}鳳梨 & 0 & 33.3 &  0 &55.6 \\
& \CJKfamily{bsmi}滑鼠 & 0 & 16.7 & 0 & 0 \\
& \CJKfamily{bsmi}紐西蘭 & 0 & 42.9 & 0 & 14.3 \\
& \CJKfamily{bsmi}印表機 & 0 & 83.3 & 0 & 50 \\
& \CJKfamily{bsmi}月台& 0 & 50 & 0 &  16.7\\
& \CJKfamily{bsmi}雲霄飛車 & 0 & 33.3 & 0 & 33.3 \\
& \CJKfamily{bsmi}鮭魚 & 0 & 0 & 0 & 60 \\
& \CJKfamily{bsmi}洗髮精 & 0 & 0 & 0 & 100 \\
& \CJKfamily{bsmi}軟體 & 0 & 30 & 0 & 20 \\
& \CJKfamily{bsmi}雪梨 & 0 & 0 & 0 & 0 \\
& \CJKfamily{bsmi}迴紋針 & 0 & 75 & 0 & 25 \\
\midrule
\multicolumn{2}{c}{\textit{Avg. PR}} &  \textbf{\textit{0}}  &  \textbf{\textit{33.1}} &  \textbf{\textit{0}} &  \textbf{\textit{34.1}} \\
\bottomrule
\end{tabular}
}
\end{CJK*}  
\caption{Each language-unique feature predominantly appears only in its own class, implying that explanation methods are capable to extract language-unique features. For example, the CN word \begin{CJK*}{UTF8}{gbsn}\textit{`菠萝'}\end{CJK*} is only used in CN's explanation, which is never used as TW's explanation, and vice versa for its translation \begin{CJK*}{UTF8}{bsmi}\textit{`鳳梨'}\end{CJK*}.}
\label{table:pr_result}
\end{center}
\vspace{-1em}
\end{table}

\subsection{PR on Low Saxon Dialects}\label{appendix:low_saxon_pr}
We conduct two additional PR experiments in Low Saxon dialects: \textit{(i)} The first and second singular pronouns in Eastphalian (OFL) are `mik/dik', compared to the rest of the Low Saxon dialects (`mi/di') \citep{siewert-etal-2020-lsdc} (Table~\ref{table:ofl_vs_non_ofl}); \textit{(ii)} The differences between all German and Dutch varieties' orthography for `house' (`Huus' vs `hoes') and `for' (`för' vs `veur') (Table~\ref{table:pr_de_vs_nl}). In both experiments, we find similar general patterns where language-unique features tend to appear in their corresponding classes, further enhancing our finding that such dialectal lexical features can be extracted by interpreting dialect classifiers.

\begin{table}[htp!]
\centering
\begin{tabular}{c|cc|cc}  
\toprule
\multicolumn{1}{c}{} & \multicolumn{2}{c}{OFL} & \multicolumn{2}{c}{Non-OFL} \\  
Count & \textit{mik} &  \textit{dik} & \textit{mi}& \textit{di} \\
\midrule
OFL Exp. & 17 & 9 & 0 & 0 \\
Non-OFL Exp. & 1 & 1 & 156 & 57 \\
\midrule
Text & 34 & 19 & 577 & 155\\ 
\midrule
OFL PR (\%) & 50.0 &47.4 & 0.0 & 0.0 \\
Non-OFL PR (\%) & 2.9 & 5.3 & 27.0 & 38.0 \\
\bottomrule
\end{tabular}
\caption{The OFL words \textit{`mik/dik'} are observed in OFL's explanations about half the time (PR=50\%), whereas the Non-OFL words \textit{`mi/di'} mainly appear in Non-OFL's explanations, indicating that explanation methods can effectively capture language-specific features.}
\label{table:ofl_vs_non_ofl}
\vspace{-1em}
\end{table}

\begin{table}[H]
\centering
\begin{tabular}{c|cc|cc}  
\toprule
 \multicolumn{1}{c}{} & \multicolumn{2}{c}{DE Feature} & \multicolumn{2}{c}{NL Feature} \\  
Count & Huus &  för & hoes& veur \\
\midrule
DE Exp. & 17 & 42 & 0 & 0 \\
NL Exp. & 0 & 0 & 0 & 23 \\
\midrule
Text & 42 & 175 & 0 & 63\\ 
\midrule
DE PR (\%) & 40.5 & 24.0 & 0.0 & 0.0 \\
NL PR (\%) & 0.0 & 0.0 & 0.0 & 36.5 \\
\bottomrule
\end{tabular}
\caption{In Low Saxon, German (DE) dialects \textit{'house'} are written \textit{'Huus'} and \textit{'for'} is written as \textit{'för'}, whereas in Dutch (NL) are written as \textit{'hoes'} and \textit{'veur'} \cite{siewert-etal-2020-lsdc}. Note that \textit{hoes} does not appear on the NL corpus. }
\label{table:pr_de_vs_nl}
\end{table}

\section{Additional Discussion}
\subsection{Why is Pick-Up Rate Low?}\label{appendix:pr_discussion}
In the CN-TW experiments, the language-unique lexical features (see Table~\ref{table:pr_result}) are always correctly identified, which means that our method has high precision. The fairly low pick-up rates, though, imply that our method has somewhat low recall. 
To test if this is indeed the case, we explore whether there exist \textit{other} lexical features that may also be language-unique but which are not part of our original feature list.
To do so, we find all features that co-occur with the features in our list for both varieties, and rank them based on their counts. We then present this updated list of possible language-unique features to our annotators (in a manner similar to \S\ref{subsec:without}). We find that the majority of them are indeed good candidates for language-unique features as well. In particular, 74.2\% of the ones selected for CN and 68.3\% of the ones for TW are indeed unique to the respective language.  This implies that the apparent low recall of our method is simply due to the presence of many good options in the data -- and the feature lists in Table~\ref{table:pr_result} contain only a subset of the possible explanations.

\subsection{Explanations from the Training Sets}\label{appendix:exp_from_train}
We explore how expanding the scale of data points influences the explanation methods' ability to capture features. 
Therefore, we run the classifier on the training set rather than the test set, to collect more explanations. We conduct experiments on LSDC dataset, similar to \sref{sec:pr_result}. Comparing the results, presented in Table~\ref{appendix:train_set_ofl_vs_non_ofl} with the original test set results in Table~\ref{table:ofl_vs_non_ofl}, we find a substantial increase in the number of data points along with noticeable fluctuations in PRs. Despite these variations, the language-specific features continue to mainly appear within their respective predicted classes. This observation reaffirms our findings that the language-unique lexical features can indeed be captured with high precision by explanations.

\begin{table}[htp!]
\centering
\begin{tabular}{c|cc|cc}  
\toprule
 \multicolumn{1}{c}{} & \multicolumn{2}{c}{OFL} & \multicolumn{2}{c}{Non-OFL} \\  
Count & \textit{mik} &  \textit{dik} & \textit{mi}& \textit{di} \\
\midrule
OFL Exp. & 92 & 44 & 0 & 0 \\
Non-OFL Exp. & 2 & 9 & 1422 & 493 \\
\midrule
Text & 263 & 150 & 4895 & 1310\\ 
\midrule
OFL PR (\%) & 35.0 & 29.3 & 0.0 & 0.0 \\
Non-OFL PR (\%) & 0.76 & 6.0 & 29.0 & 38.0 \\
\bottomrule
\end{tabular}
\caption{Despite an increase in data points and fluctuations in PR values, language-specific features consistently appear predominantly within their respective predicted classes. This observation strengthens our previous findings, reaffirming that the explanation method can indeed effectively capture language-unique features, regardless of the scale of data points.}
\label{appendix:train_set_ofl_vs_non_ofl}
\end{table}

\begin{table}[H]
\centering
\begin{tabular}{c|cc|cc}  
\toprule
 \multicolumn{1}{c}{} & \multicolumn{2}{c}{DE Feature} & \multicolumn{2}{c}{NL Feature} \\  
Count & Huus &  för & hoes& veur \\
\midrule
DE Exp. & 101 & 388 & 3 & 0 \\
NL Exp. & 2 & 0 & 0 & 182 \\
\midrule
Text & 327 & 1492 & 5 & 559\\ 
\midrule
DE PR (\%) & 30.1 & 26.0 & 6.0 & 0.0 \\
NL PR (\%) & 0.6 & 0.0 & 0.0 & 33.0 \\
\bottomrule
\end{tabular}
\caption{Evaluating explanations from the training sets for DE vs NL.}
\label{appendix:train_as_test_de_vs_nl}
\end{table}

\subsection{Other Features}\label{appendix:other_features}
While our primary focus is extracting lexical features from dialects, we also explore extracting sub-word features. We conducted a study using the LSDC dataset \cite{siewert-etal-2020-lsdc}, which contains examples where the plural suffix of verbs in the present tense differs among dialects. In dialects MKB, MAR, NPR, OFR, and GRO (class 1), the plural suffix is \textit{-(e)n}, while in the rest of the dialects in the dataset (class 0), it is \textit{-(e)t}. We counted the occurrences of these two suffixes in the text and used PR to evaluate whether explanation models can recognize these subtle, language-specific features. Table~\ref{table:en_vs_et} illustrates the results. For class 1 the models do indeed return features with its unique feature \textit{-(e)n} with a relatively high PR (20\%) and it (correctly) does not return features for class 0 (only 1\% PR).

The other type of ending (\textit{-(e)t}), on the other hand, is not returned as part of the model explanations for class 0. We hypothesize that this discrepancy is due to feature overload: several other words in these dialects, which are not present-tense verbs, have the same ending. To accurately capture these sub-word features, further investigation is necessary, along with the development of morphological and morphosyntactic analysis tools for these dialects, which extends beyond the scope of our current work.

\begin{table}[htp!]
\centering
\resizebox{0.35\textwidth}{!}{%
\begin{tabular}{c|c|c}  
\toprule
Count & \textit{-et} (C.0) &  \textit{-en} (C.1) \\
\midrule
C.0 Exp. & 39 & 209   \\
C.1 Exp. & 904  & 4046  \\
\midrule

Text  & 3772 & 21112\\ 
\midrule
C.0 PR (\%) & 1.0 & 1.0  \\
C.1 PR (\%) & 24.0  & 19.2  \\
\bottomrule
\end{tabular}
}
\caption{The PR for the class 1 unique feature \textit{-en} is higher within its class, but the model's recognition of the subtle morphological distinction is unclear, given the PR for \textit{-et} is 1\% within its class. This could be attributed to the inclusion of non-present-tense verbs with \textit{-et} endings.}
\label{table:en_vs_et}
\end{table}


\clearpage
\newpage
\section{Classification Results}\label{appendix:classification_acc_baseline}
\begin{table}[th!]
\centering
\begin{tabular}{ccc|c}
\toprule
{} & Dialect & \multicolumn{2}{c}{Accuracy} \\
\multirow{8}{*}{Italian} &  &Baseline & Ours \\ 
\hline
{} & EML & 98.3 &  99.1\\ 
{} & FUR & 99.7 & 99.8\\ 
{} & LIJ & 99.7 & 100.0\\ 
{} & LMO & 99.7 & 99.3\\ 
{} & NAP & 100.0 & 99.2\\ 
{} & PMS & 100.0 & 99.7\\ 
{} & ROA & 100.0 & 100.0\\ 
{} & SC & 98.5 & 97.4\\ 
{} & SCN & 99.7 & 99.2\\ 
{} & VEC & 99.7 & 99.4\\ 
\hline
\multirow{7}{*}{Low Saxon} & ACH & 93.6 & 99.1\\ 
{} & DRE & 97.7 & 98.6\\ 
{} & HAM & 94.1& 95.0\\ 
{} & HOL & 96.0 & 93.8\\ 
{} & MAR & 83.3 & 99.8\\ 
{} & MKB & 96.5 & 96.8\\ 
{} & MON & 85.7 & 99.1\\ 
{} & NPR & 87.0 & 99.6\\ 
{} & OFL & 97.7 & 98.3\\ 
{} & OVY & 87.0 & 99.5\\ 
\hline
\multirow{1}{*}{Mandarin} & TW & 99.3 & 99.4 \\
 \hline
\end{tabular}
\caption{Pairwise test set accuracy of baseline classifiers versus interpretable classifiers. Our method achieve generally high accuracy for all language pairs. We see equally high performance on Italian and Mandarin dialects. However we see a disparity in performance across Low Saxon dialects. }
\end{table}


\section{Feature Counts for CN-TW PR}\label{appendix:word_counts}
\begin{figure}[H]
    \centering  
    \includegraphics[width=7.5cm]{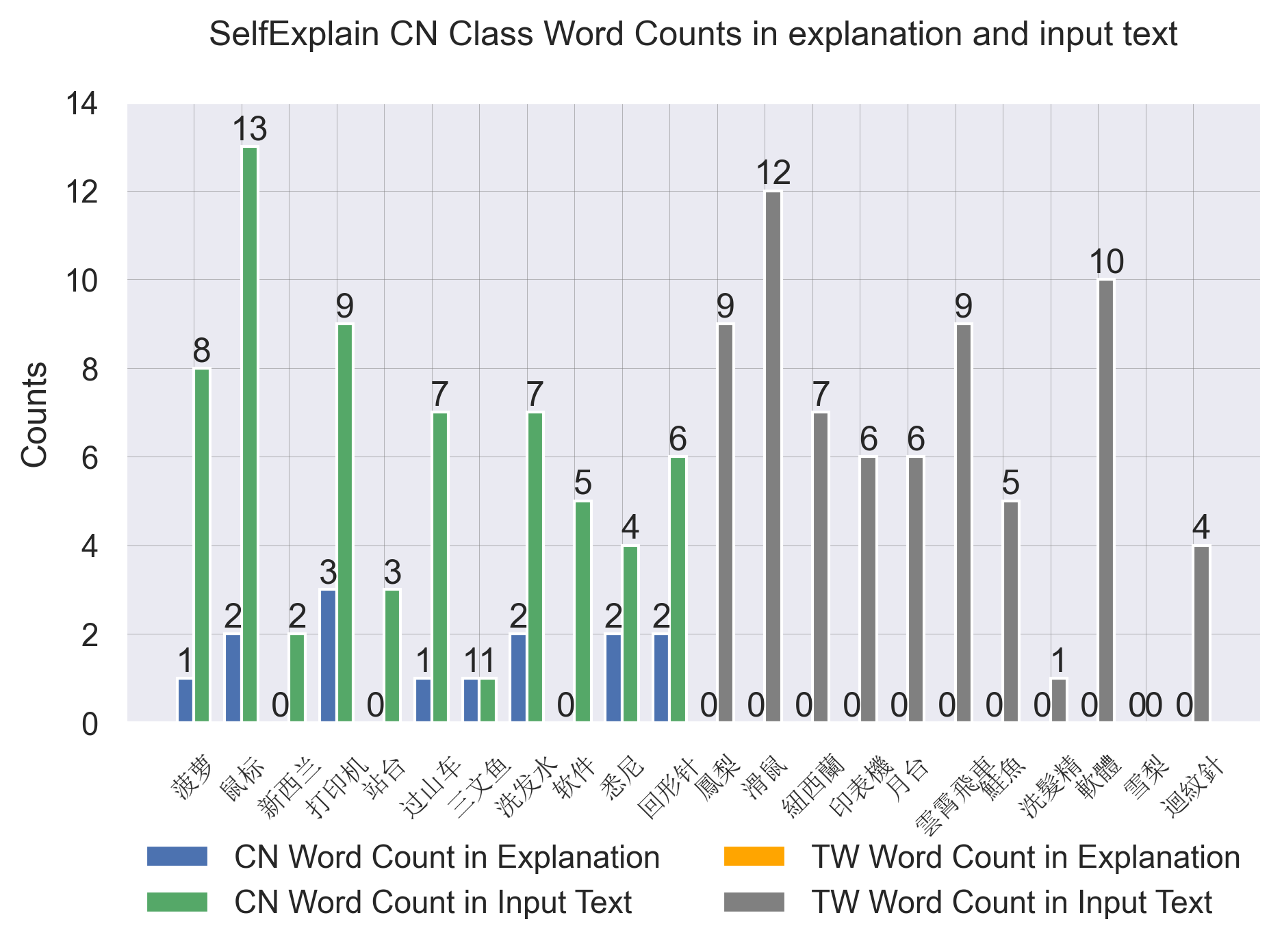}
        \caption{SelfExplain CN Class feature counts in explanation and input text. }

    \label{appendix:selfexp_cn_word_counts}
\end{figure}

\begin{figure}[H]
    \centering  
    \includegraphics[width=7.5cm]{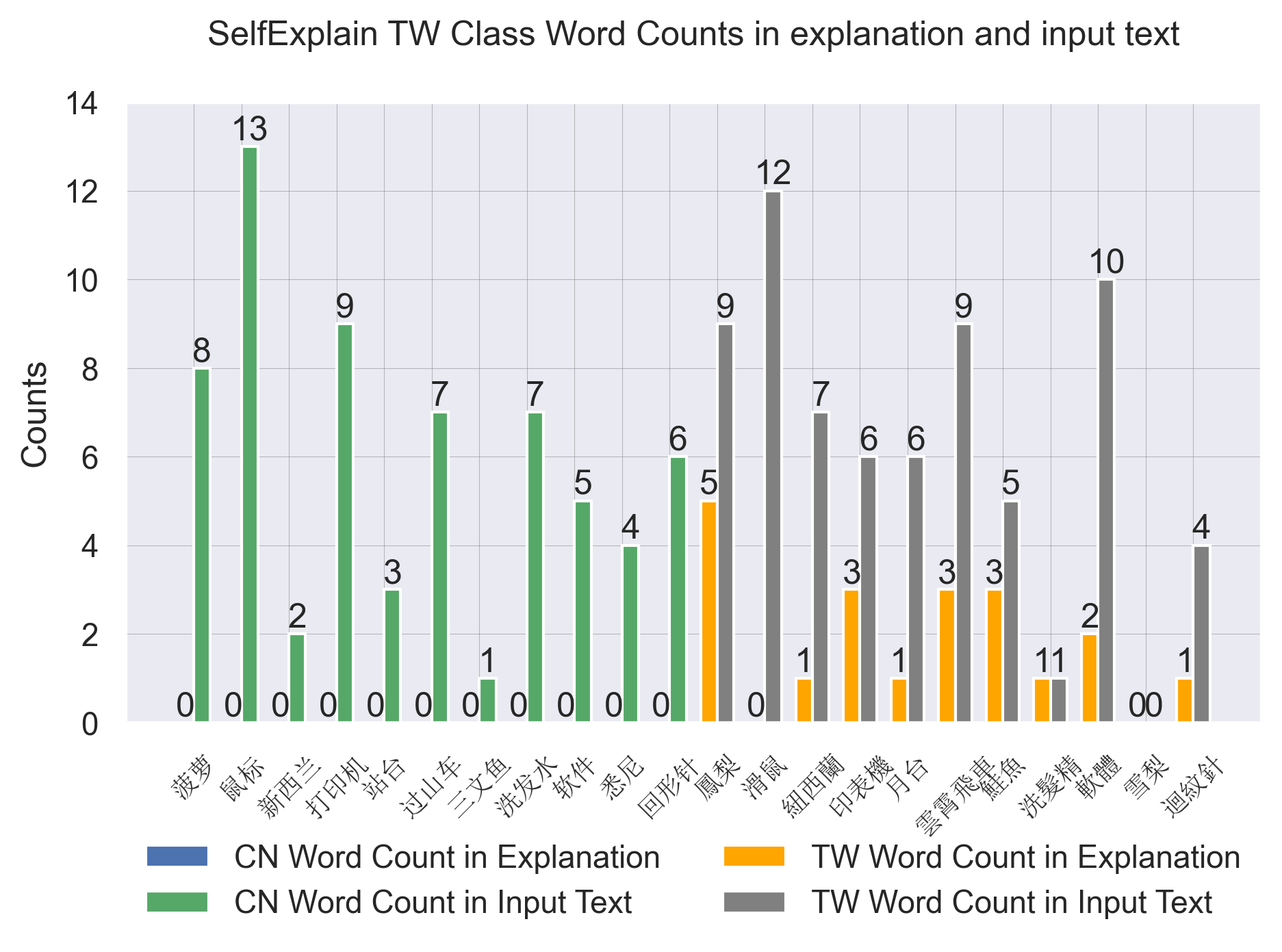}
        \caption{SelfExplain TW Class feature counts in explanation and input text.}

    \label{appendix:selfexp_tw_word_counts}
\end{figure}

\begin{figure}[H]
    \centering  
    \includegraphics[width=7.5cm]{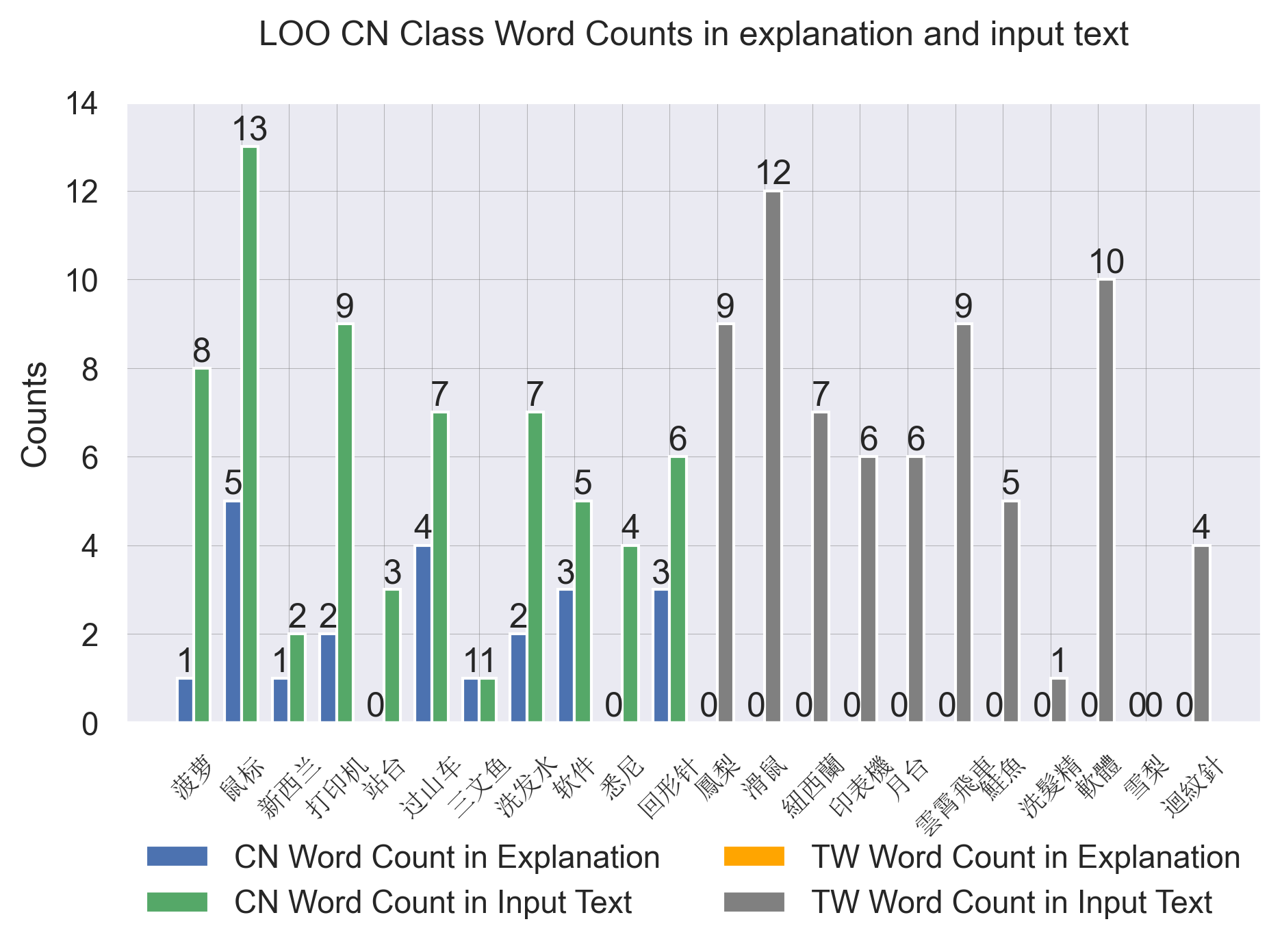}
        \caption{LOO CN Class feature counts in explanation and input text.}

    \label{appendix:loo_cn_class_word_counts}
\end{figure}

\begin{figure}[H]
    \centering  
    \includegraphics[width=7.5cm]{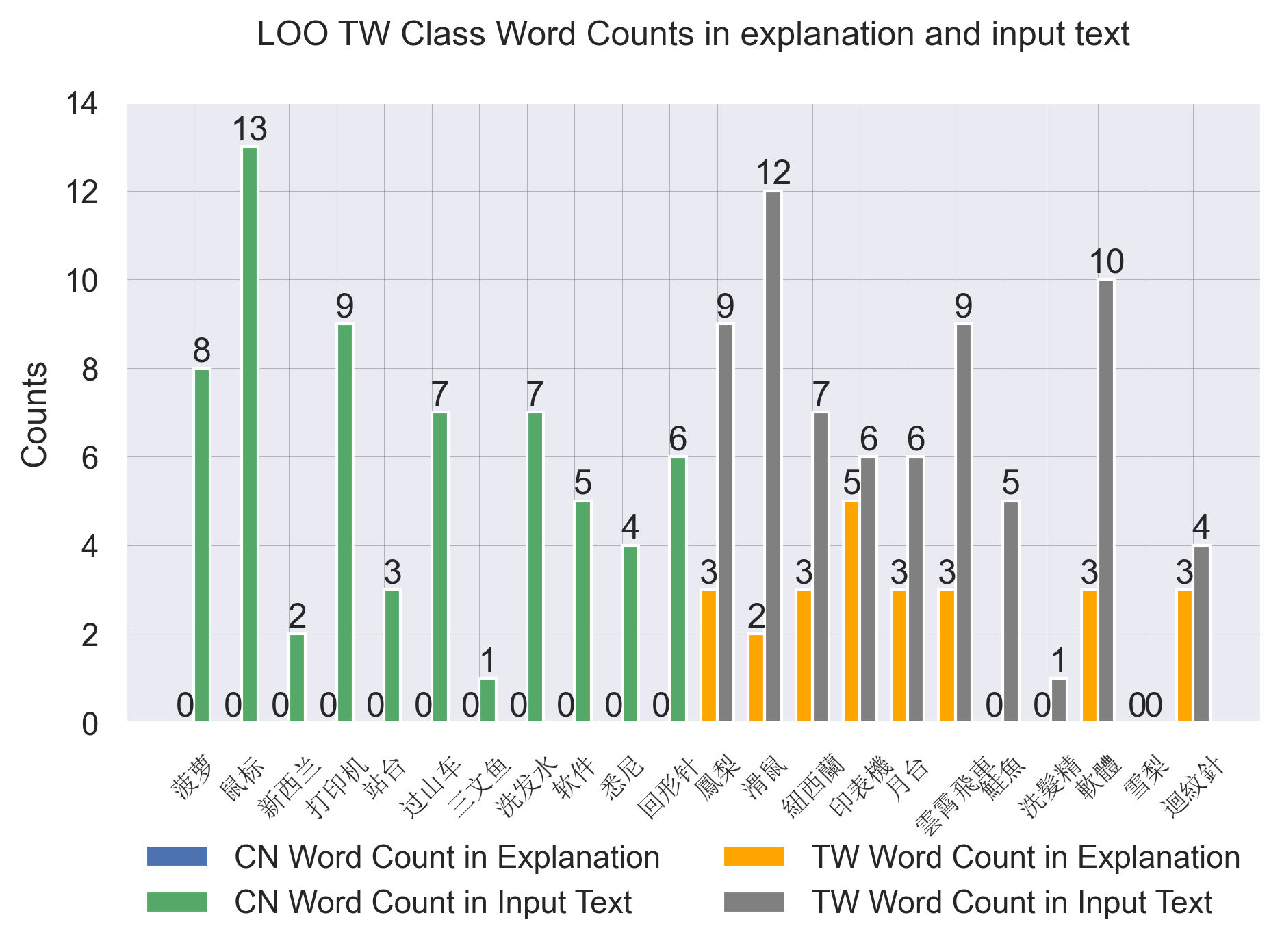}
        \caption{LOO TW Class feature counts in explanation and input text.}

    \label{appendix:loo_tw_class_word_counts}
\end{figure}

\section{Inter-annotator Agreement Statistics}\label{appendix:agreement_statistics}
To minimize potential biases, we mixed the features between both classes.

\begin{figure*}[htbp]
    \centering
    \includegraphics[width=\textwidth]{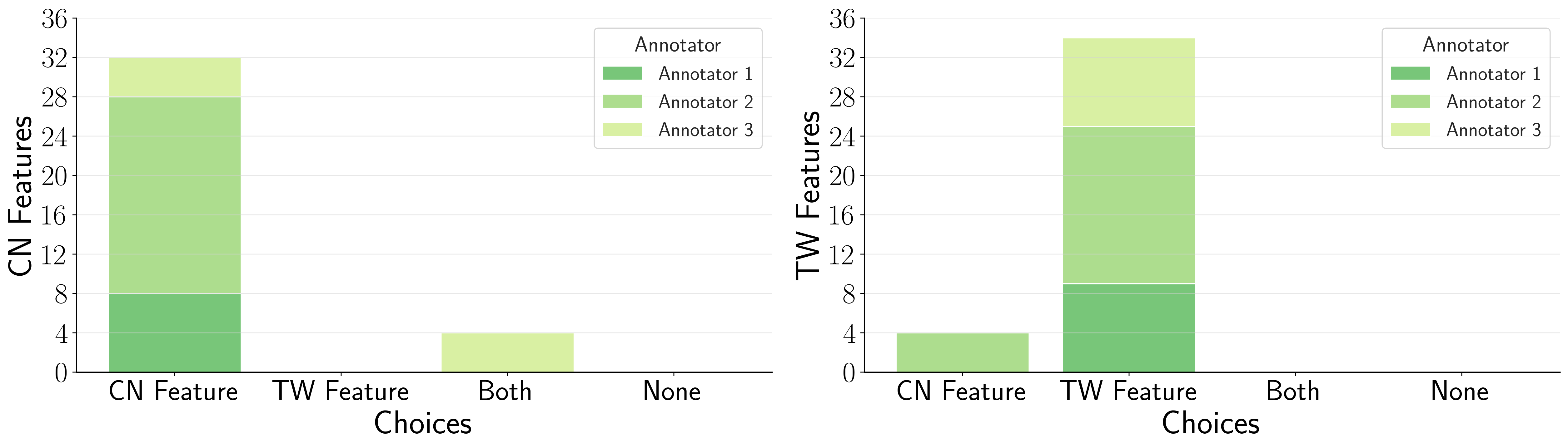}
    \caption{Inter-annotator agreement statistics on extracted CN features (left) and TW features (right). Most extracted features align with human annotators.}
    \label{appendix:combined_annotator_agreement_CN_TW}
\end{figure*}

\begin{figure}[H]
    \centering  
    \includegraphics[width=\textwidth]{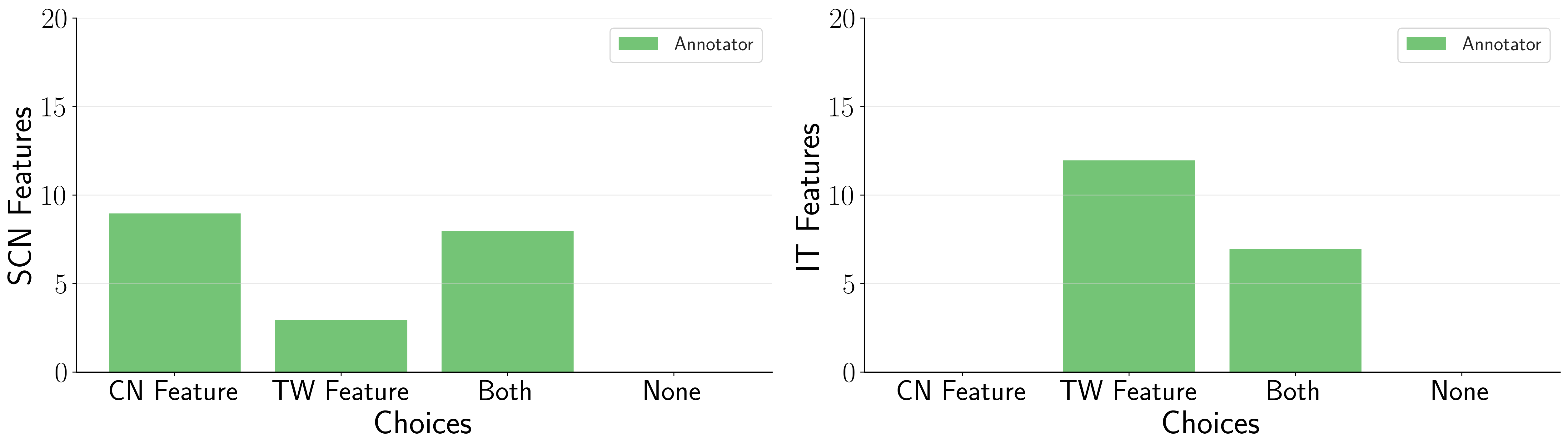}
        \caption{Inter-annotator agreement statistics on SCN (left) and IT (right) features. Note that there is only one annotator in SCN-IT experiment.}

    \label{appendix:combined_annotator_agreement_SCN_IT}
\end{figure}

\end{document}